\documentclass[journal]{IEEEtran}
\ifCLASSINFOpdf
\else
\fi

\usepackage{bm}
\usepackage{url}
\usepackage{graphicx}
\usepackage{lipsum}
\usepackage{amsmath}
\usepackage{amssymb}
\usepackage{nccmath}
\usepackage{comment}
\usepackage{balance}
\usepackage{multirow}
\usepackage{textcomp}
\usepackage{multirow}
\usepackage{subcaption}
\usepackage{algorithm}
\usepackage{dblfloatfix}
\usepackage[table]{xcolor}
\usepackage[noend]{algpseudocode}

\begin{document}

%
\title{Globally Consistent 3D LiDAR Mapping with GPU-accelerated GICP Matching Cost Factors}
%
%
%

\author{Kenji Koide$^{1}$, Masashi Yokozuka$^{1}$, Shuji Oishi$^{1}$, and Atsuhiko Banno$^{1}$
\thanks{*This work was supported in part by a project commissioned by the New Energy and Industrial Technology Development Organization (NEDO).}
\thanks{$^{1}$All the authors are with the Department of Information Technology and Human Factors, the National Institute of Advanced Industrial Science and Technology, Umezono 1-1-1, Tsukuba, 3050061, Ibaraki, Japan, {\tt\small k.koide@aist.go.jp}}%
}

%
%

\markboth{IEEE ROBOTICS AND AUTOMATION LETTERS. PREPRINT VERSION. ACCEPTED SEPTEMBER, 2021}%
{Shell \MakeLowercase{\textit{et al.}}: Bare Demo of IEEEtran.cls for IEEE Journals}
%



\makeatletter
\g@addto@macro\@maketitle{
  \begin{figure}[H]
  \setlength{\linewidth}{\textwidth}
  \setlength{\hsize}{\textwidth}
  \centering
  \includegraphics[width=1.0\linewidth]{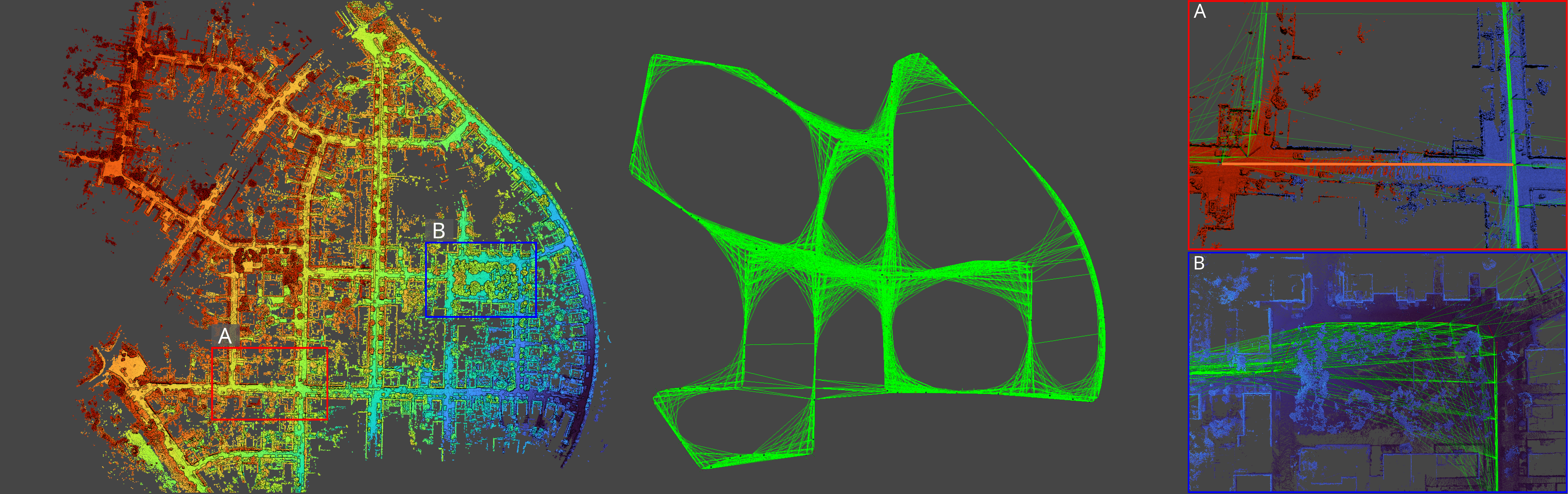}
  \caption{Mapping result for the KITTI 00 sequence. (A) The global matching cost minimization approach enables constraint of the relative pose between frames with a small overlap, and (B) the GPU-powered implementation allows creation of a massive amount of factors and the construction of a densely connected factor graph. The factor graph contains over 4,500 matching cost factors, and each factor involves the cost evaluation of approximately 20,000 points on average. The optimization converges in a few seconds on a middle-class GPU.}
  \label{fig:kitti00}
  \vspace{-5mm}
  \end{figure}
}
\makeatother

\setcounter{figure}{-2}
\renewcommand{\thesubfigure}{\Alph{subfigure}}

\maketitle

\begin{abstract}
This paper presents a real-time 3D LiDAR mapping framework based on global matching cost minimization. The proposed method constructs a factor graph that directly minimizes matching costs between frames over the entire map, unlike pose graph-based approaches that minimize errors in the pose space. For real-time global matching cost minimization, we use a voxel data association-based GICP matching cost factor that is able to fully leverage GPU parallel processing. The combination of the matching cost factor and GPU computation enables constraint of the relative pose between frames with a small overlap and creation of a densely connected factor graph. The mapping process is managed based on a voxel-based overlap metric that can quickly be evaluated on a GPU. We incorporate the proposed method with an external loop detection method in order to help the voxel-based matching cost factors to avoid convergence in a local solution. The experimental result on the KITTI dataset shows that the proposed approach improves the estimation accuracy of long trajectories.
\end{abstract}


%
\IEEEpeerreviewmaketitle

\section{Introduction}

\IEEEPARstart{E}{nvironmental} mapping is crucial for autonomous systems, and SLAM has been a major research topic in the robotics field. An important aspect of SLAM is global consistency. It is desirable that a mapping system is able to retain the consistency of every single part of a map after running on a long trajectory and closing large loops.

One way to refine a trajectory estimation result and improve the mapping consistency is pose graph optimization, which minimizes the relative pose errors between frames in the pose space \cite{Grisetti2010}. This approach has been well established in the literature and is widely used \cite{Behley2018, Reijgwart2020}. Pose graph optimization requires modeling each relative pose constraint in the form of a Gaussian distribution (i.e., mean and covariance matrix). However, representing a relative pose, which is typically a result of scan matching, as a Gaussian distribution is obviously too approximated. Scan matching solutions have many local minima and thus cannot be accurately modeled in the form of a unimodal distribution. Furthermore, estimating the uncertainty (i.e., covariance matrix) of a scan matching result is difficult in practice \cite{Landry2019}. Most existing studies use only a constant covariance matrix \cite{Behley2018}, a simple weighting scheme \cite{liosam2020shan}, or Hessian-based closed-form covariance estimation \cite{Hess2016}, which tends to be optimistic \cite{Landry2019}. Inaccurate modeling of relative pose constraints can lead to deteriorated estimation accuracy of a long trajectory with large loops.

Global matching cost minimization is another approach to improve the consistency of a map. Early on, Lu and Milios proposed a graph-based 2D mapping approach that minimized the matching cost between frames over the entire map \cite{Lu1997}. This method was then extended to three dimensions by reducing the number of global optimization executions by explicitly handling loop closure events \cite{Borrmann2008, Sprickerhof2011}. These approaches ensured that all the frames were aligned together and thereby retained the local consistency of every part of the map (i.e., global consistency) while closing loops. They evaluated the global matching cost and updated each factor for every optimization iteration; this can be interpreted as the SE3 relative pose factor with variable mean and covariance. Furthermore, each factor can represent a deficient constraint in this way. They thus can more accurately model the constraint of the relative pose between frames compared to pose graph-based approaches. However, performing global matching cost minimization was still computationally expensive, and application to large-scale and real-time mapping was considered to be infeasible.

In this work, we revisit the global matching cost minimization approach with modern GPU computation techniques and propose a real-time and globally consistent 3D LiDAR mapping framework. The core of the proposed framework is the multi-scan registration algorithm, which minimizes the errors of Generalized ICP (GICP) matching cost factors with voxel-based data association \cite{vgicp,Segal2009} over the entire map by fully leveraging GPU parallel processing. This approach has several advantages. First, this enables constraint of the relative pose between frames with a very small overlap, where it is difficult to explicitly estimate the relative pose through scan matching (Fig. \ref{fig:kitti00}(A)). Second, the GPU-powered implementation enables the creation of a massive amount of factors (Fig. \ref{fig:kitti00}(B)). Although we create a matching cost factor between every frame pair with an overlap rate larger than a small threshold (e.g., 2.5\%), the global map optimization converges in a few seconds on a middle-class GPU.

The proposed framework consists of local and global mapping modules, which perform matching cost minimization locally and globally, respectively (see Fig. \ref{fig:system}). Both of the mapping modules are managed based on a voxel-based overlap metric that can quickly be evaluated on a GPU. In order to prevent the voxel-based matching cost factors from becoming stuck at a local minimum, we explicitly detect a few loops with an external loop detector (e.g., ScanContext \cite{Kim2018}) and add these loops to the factor graph as SE3 relative pose constraints. Through evaluation on the KITTI dataset \cite{Geiger2012}, we show that the proposed approach improves the estimation accuracy of long trajectories with large loops.

The key contributions of this work are as follows:
\begin{enumerate}
  \item We present a globally consistent 3D mapping framework based on the GPU-accelerated matching cost factor and show that the matching cost minimization over the entire map is feasible in real-time. To the best of our knowledge, this is the first real-time method that performs scan matching at a global scale.
  \item We propose a mapping management mechanism based on an overlap metric that can quickly be evaluated on a GPU and enables the design of a general mapping process.
  \item We show that the global matching cost minimization approach enables retention of the global consistency of large maps and increases the mapping quality.
\end{enumerate}

\section{Related Work}

Since the main contribution of this work is a method by which to retain the global consistency of a map, we focus in this section on global map optimization approaches.

\subsection{Pose Graph Optimization}


While many frontend methods for 3D LiDAR SLAM have been proposed \cite{Behley2018, liosam2020shan, yokozuka2021, pan2021mulls}, most of these systems rely on pose graph-based maximum a posteriori estimation \cite{Grisetti2010} as the backend in order to refine trajectory estimation results and improve the mapping consistency. Pose graph-based approaches construct a factor graph with relative pose (SE3) constraints and estimate the sensor trajectory that minimizes the errors in the pose space. This approach has been well established and has become the gold standard for the 3D LiDAR SLAM backend.

In pose graph optimization, relative pose constraints are modeled as a Gaussian distribution. However, the Gaussian distribution form is a too-approximated representation for scan matching results. A scan matching solution inherently has many local minima and thus cannot be accurately modeled in the unimodal distribution form, and this approximated representation would affect the optimization result once the scan matching converges to a local solution.

Furthermore, estimation of the covariance matrix of a scan matching result is difficult in practice \cite{Landry2019}. Although closed-form uncertainty estimation methods based on the Hessian matrix of the cost function have been commonly used \cite{Bengtsson2003,Hess2016}, it is known that closed-form methods tend to be optimistic because these methods are not able to take into account the cost function deviation caused by data association changes \cite{Landry2019}. On the other hand, Monte-Carlo-based covariance estimation methods can more accurately estimate the uncertainty of scan matching results \cite{Iversen2017}. These Monte-Carlo-based methods, however, are computationally expensive. Although data-driven covariance estimation approaches \cite{Landry2019} have been proposed in order to balance the real-time performance and estimation accuracy, most existing SLAM frameworks use only a constant covariance matrix \cite{Behley2018}, a simple weighting scheme \cite{liosam2020shan, yokozuka2021}, or Hessian-based closed form covariance estimation \cite{Hess2016}.

\subsection{Deformation Graph}

Map deformation is another approach to retain the surface consistency of mapping results including loops \cite{Whelan2013}. This approach constructs a graph that deforms the mapping result such that the local consistency is preserved. Deformation graph-based map-centric mapping approaches without estimation of the full sensor trajectory have been proposed \cite{Park2018} . These methods, however, do not accurately take all available information into account and may disrupt the global consistency of the map.

\subsection{Bundle Adjustment}

Bundle adjustment (BA) that simultaneously optimizes sensor poses and environmental parameters over frames has been important in the visual SLAM field \cite{MurArtal2017}. It has been shown that BA-based methods show good trajectory estimation and reconstruction accuracy while it is known to be computationally expensive. For real-time performance, BA is typically carried out at two different scale levels (real-time local BA and low frequency global BA) with a limited number of feature points  \cite{MurArtal2017}. Notably, Sch\"{o}ps {\it et al.} recently showed that direct BA-based visual odometry at a global level is feasible in real-time with GPU processing \cite{Schops2019}. In the context of 3D LiDAR SLAM, however, it is rare to see BA-based approaches due to the difficulty of feature tracking on sparse LiDAR data, and few studies on BA-based approaches have been proposed \cite{liu2020balm,Wisth2021}.

\subsection{Global Matching Cost Minimization}

Lu and Milios formulated the mapping problem as minimization of the matching cost of frames over a factor graph \cite{Lu1997}, and their method was extended to three dimensions by several studies by explicitly handling loop detection events \cite{Borrmann2008,Sprickerhof2011}, their method was still considered to be infeasible in real-time because it needs to re-evaluate the matching cost of every frame pair at every optimization iteration. 

Recently, Reijgwart {\it et al.} proposed a volumetric mapping method that takes into account registration errors between local submaps \cite{Reijgwart2020}. They used an efficient registration error metric based on Euclidean Signed Distance Field (ESDF) representation in order to avoid the costly correspondence search. The cost evaluation was, however, still computationally expensive, and optimization was carried out with only a random subset of residuals and with the support of SE3 relative pose constraints.

\subsection{GPU-based SLAM}

The GPU has been commonly used for almost every component of dense visual and RGB-D SLAM (from frontend \cite{Schops2019} to backend \cite{Whelan2013}). In contrast, in the context of LiDAR SLAM, the use of GPU was mostly limited to accelerating scan matching in the frontend \cite{Behley2018, elo}. While there have been also proposed deep learning-based frontend \cite{Li2019} and loop detection methods \cite{Chen2020} with GPU processing, in most of the works, pose graph optimization performed on a CPU is in charge of global optimization.

\begin{figure}[tb]
  \centering
  \includegraphics[width=0.85\linewidth]{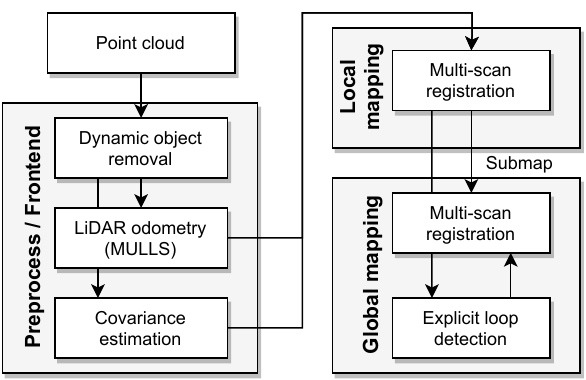}
  \caption{Overview of the proposed framework.}
  \label{fig:system}
\end{figure}

\section{Methodology}

\begin{figure}[tb]
  \centering
  \includegraphics[width=1.0\linewidth]{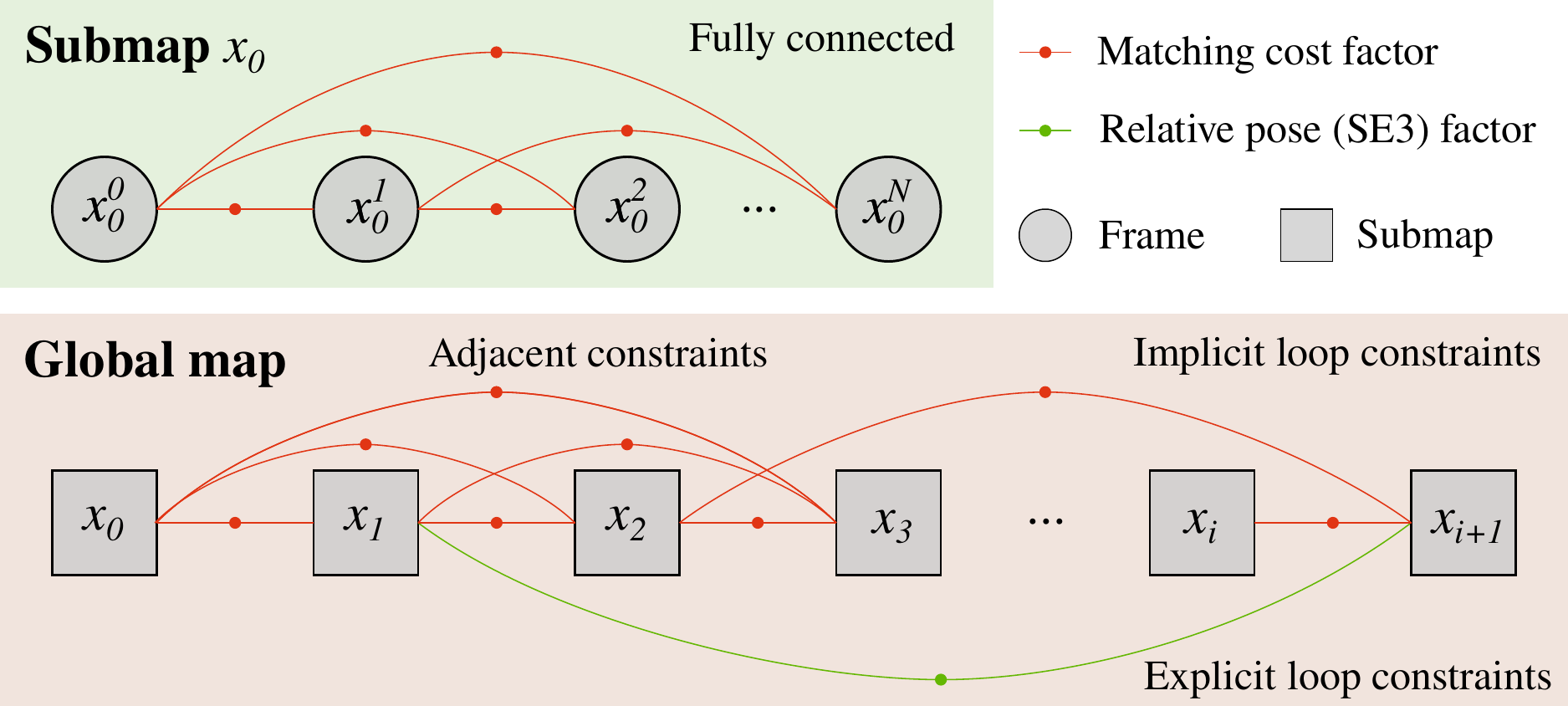}
  \caption{Factor graph of the proposed framework. For local submapping, a fully connected matching cost factor graph is constructed. For global mapping, matching cost factors are created between every keyframe pair with an overlap rate larger than a threshold. In order to prevent the matching cost factors from becoming trapped at a local solution, explicit SE3 loop constraints are inserted into the graph. }
  \label{fig:graph}
\end{figure}

Figure \ref{fig:system} shows an overview of the proposed system. We first remove dynamic objects (e.g., cars and pedestrians) from input point clouds using RandLA-Net \cite{Hu2020} and run an odometry estimation algorithm (e.g., MULLS \cite{pan2021mulls}) to obtain an initial guess for the latest sensor pose. Meanwhile, we estimate the covariance matrix of each point from its k-neighboring points. Note that the costly nearest neighbor search is used only in this preprocessing step, which is performed once for every input point cloud. 

The preprocessed point cloud and the sensor pose initial guess are fed to the local mapping module that merges approximately 10 to 20 frames into one submap, and the submaps are then merged into one global map in the following global mapping module. The core of the local and global mapping modules is the multi-scan registration algorithm that constructs a factor graph with voxelized GICP matching cost factors (see Fig. \ref{fig:graph}). This optimizes the sensor poses such that all neighboring frames are aligned together. In the submapping module, we construct a fully-connected factor graph. In the global-mapping module, we create constraints between the latest submap and every past submap that has a certain overlap with the latest submap. As a result, all of the submaps are aligned with not only adjacent submaps on the graph but also every revisited submap that results in closing loops implicitly. We obtain the final mapping result by concatenating submap point clouds based on the optimized trajectory.


\subsection {Voxelized GICP Matching Cost Factor}

We estimate a set of sensor poses $\mathcal{T} = \{ {\bm T}_0, \cdots, {\bm T}_t \}$ by minimizing the matching cost over a set of point cloud pairs $\mathcal{F}^M$. The objective function to be minimized is defined as:

\begin{align}
\label{eq:objective}
f^M(\mathcal{F}^M, \mathcal{T}) = \sum_{(i, j) \in \mathcal{F}^M} e^M(\mathcal{P}_i, \mathcal{P}_j, {\bm T}_i, {\bm T}_j),
\end{align}
where $\mathcal{P}_i$ and $\mathcal{P}_j$ are a point cloud pair, ${\bm T}_i$ and ${\bm T}_j$ are their poses, and $e^M$ is a matching cost function.

As the matching cost function, we choose the voxelized GICP (VGICP) cost \cite{vgicp} that is as accurate as GICP and suitable for GPU processing. The VGICP cost is based on the GICP distribution-to-distribution error that is the most accurate among the ICP variants \cite{Segal2009}. The GICP error between a point with covariance $\bm{p}_k = (\bm{\mu}_k, \bm{C}_k)$ and its corresponding point $\bm{p}'_k = (\bm{\mu}'_k, \bm{C}'_k)$ on a transformation $\bm{T}$ is defined as:
\begin{align}
\label{eq:gicp}
e^{\text{\it GICP}}(\bm{p}_k, \bm{T}) = \bm{d}_k^T (\bm{C}'_k + \bm{T} \bm{C}_k \bm{T}^T) ^{-1} \bm{d}_k,
\end{align}
where $\bm{d}_k = \bm{\mu}'_k - \bm{T} \bm{\mu}_k$ is the residual between $\bm{\mu}_k$ and $\bm{\mu}_k'$.

In the original GICP algorithm, corresponding points are given by a nearest neighbor search, e.g., by a KD tree. However, the use of a KD tree is not suitable for a GPU because the KD tree uses a number of conditional branches, which affects the performance of the GPU. In order to maximize the processing speed, VGICP uses a voxel-based data association approach. It discretizes each input point cloud into voxels at resolution $r$ and calculates the mean and covariance of each voxel based on the points that fall within the voxel. VGICP aggregates point distributions into one voxel distribution, unlike Normal Distributions Transform (NDT)-based algorithms that compute a voxel distribution from a set of points \cite{Stoyanov2012}. This approach enables a valid distribution on a voxel to be obtained with only a few points and results in robustness to voxel resolution changes and more accuracy than NDT \cite{vgicp}.

Then, the matching cost between a point cloud $\mathcal{P}_i = \{ \bm{p}_0, \cdots, \bm{p}_N \}$ and another point cloud $\mathcal{P}_j$ is defined as:
\begin{align}
\label{eq:cost}
e^M(\mathcal{P}_i, \mathcal{P}_j, \bm{T}_i, \bm{T}_j) = \sum_{\bm{p}_k \in \mathcal{P}_i} e^{\text{\it GICP}}(\bm{p}_k, \bm{T}_{ij}),
\end{align}
where $\bm{T}_{ij} = \bm{T}_i^{-1} \bm{T}_j$ is the relative pose estimate between $\mathcal{P}_i$ and $\mathcal{P}_j$. The corresponding points $\bm{p}'_k$ are given by looking up the voxel map of $\mathcal{P}_j$. From the derivatives of Eq. \ref{eq:cost}, we obtain a Hessian factor to constrain $\bm{T}_i$ and $\bm{T}_j$ that is composed of Hessian matrices $\bm{H}_{ii}, \bm{H}_{ij}$, and $\bm{H}_{jj}$ and coefficient vectors $\bm{b}_i$ and $\bm{b}_j$:

\begin{align}
\bm{A}_k &= \frac{\partial \bm{e}_k}{\partial \bm{T}_i}, \ 
\bm{B}_k = \frac{\partial \bm{e}_k}{\partial \bm{T}_j}, \\
\nonumber
\bm{H}_{ii} &= \sum_k^N \bm{A}^T_k \bm{\Omega}_k \bm{A}_k, \ 
\bm{H}_{ij}  = \sum_k^N \bm{A}^T_k \bm{\Omega}_k \bm{B}_k, \\
\label{eq:hessian}
\bm{H}_{jj} &= \sum_k^N \bm{B}^T_k \bm{\Omega}_k \bm{B}_k, \\
\label{eq:coefficients}
\bm{b}_i &= \sum_k^N \bm{A}_k^T \bm{\Omega}_k \bm{e}_k, \ 
\bm{b}_j  = \sum_k^N \bm{B}_k^T \bm{\Omega}_k \bm{e}_k,
\end{align}
where $\bm{e}_k = \bm{\mu}'_k - \bm{T}_{ij} \bm{\mu}_k$, and $\bm{\Omega}_k = \left( \bm{C}'_k + \bm{T}_{ij} \bm{C}_k \bm{T}_{ij}^T \right)^{-1}$.
Note that we re-evaluate the matching cost function $e^M$ every optimization iteration, and thus $\bm{H}_*$ and $\bm{b}_*$ are also updated at the current linearization point.

\subsection {Local Mapping}

The local mapping module aggregates a number of consecutive frames into one local submap in order to reduce the number of pose variables optimized in the following global mapping module.

In order to manage the mapping process, we use criteria based on a simple fine-grained voxel-based overlap metric. We define the overlap rate between two point clouds $\mathcal{P}_i$ and $\mathcal{P}_j$ as the fraction of points $\bm{p}_k \in \mathcal{P}_i$ that fall within a voxel of $\mathcal{P}_j$:
\begin{align}
s(\bm{p_k}, \mathcal{P}_j) &= 
  \begin{cases}
    1 & \text{if $\bm{p}_k$ fell in a voxel of $\mathcal{P}_j$} \\
    0 & \text{otherwise}
  \end{cases} \\
\label{eq:overlap}
\mbox{overlap}(\mathcal{P}_i, \mathcal{P}_j) &= \frac{\sum_k^N s(\bm{p}_k, \mathcal{P}_j)}{N}.
\end{align}
Equation \ref{eq:overlap} can quickly be evaluated on a GPU, and evaluation takes less than 0.1 ms for a point cloud pair with approximately 50,000 points for each. This voxel-based overlap metric enables the design of a general mapping process compared to metrics based on time interval \cite{liosam2020shan} or sensor displacement \cite{pan2021mulls} that require careful tuning of parameters depending on the environment, while more accurately detecting overlapping frames as compared to bounding box-based overlap metrics \cite{Reijgwart2020}.

If the overlap between the current frame and the last frame in the submap is larger than a threshold $\mbox{\it th}^L_{\text{\it max}}$ (e.g., 95\%), then the sensor is considered not to have made a move, and we skip that frame. Otherwise, we create its voxel map with resolution of $r^L$ and insert the pair of the current frame and the voxel map into the submap factor graph. We create matching cost factors between the inserted frame and all the other frames in the submap, and thus a fully connected factor graph is created for local mapping. Whenever the overlap between the very first and last frames in the submap becomes smaller than threshold $\text{\it th}^L_{\text{\it min}}$ (e.g., 10\%) or the number of frames in the factor graph becomes larger than threshold $N^L_{\text{\it max}}$, we perform factor graph optimization and merge all of the frames into one submap based on the optimized sensor poses. A voxel map with a resolution of $r^G$ is created from the submap and then fed to the following global mapping module. We assume that the estimation drift in the short time span of the submap window is negligible and fix the relative poses between frames in the submap in the global mapping. 

\subsection {Global Mapping}

\begin{figure}[t]
  \centering
  \begin{minipage}[b]{0.48\linewidth}
  \centering
  \includegraphics[height=4.0cm]{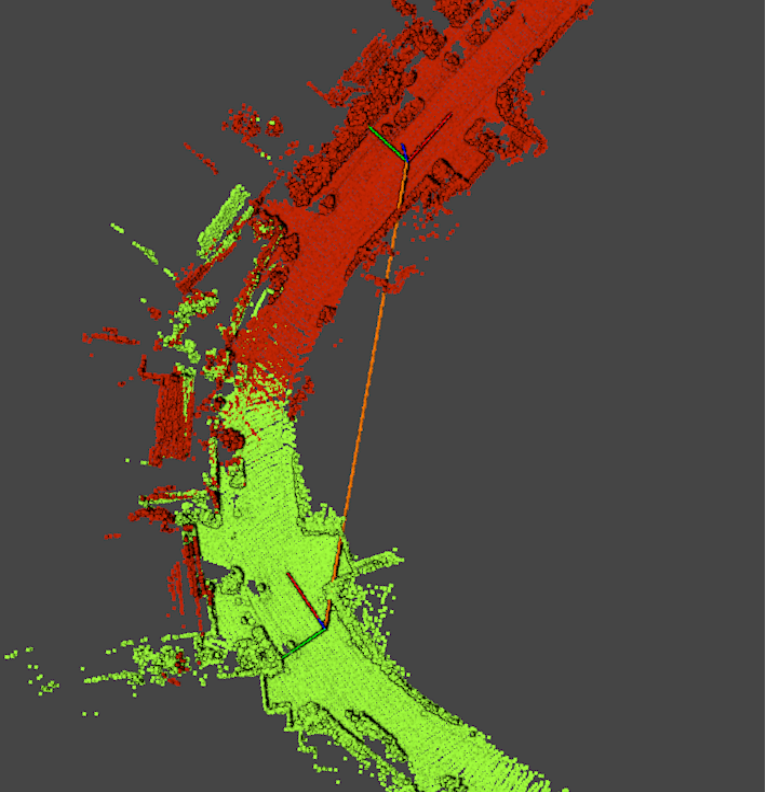}
  \subcaption{Matching cost factor}
  \end{minipage}
  \begin{minipage}[b]{0.48\linewidth}
  \centering
  \includegraphics[height=4.0cm]{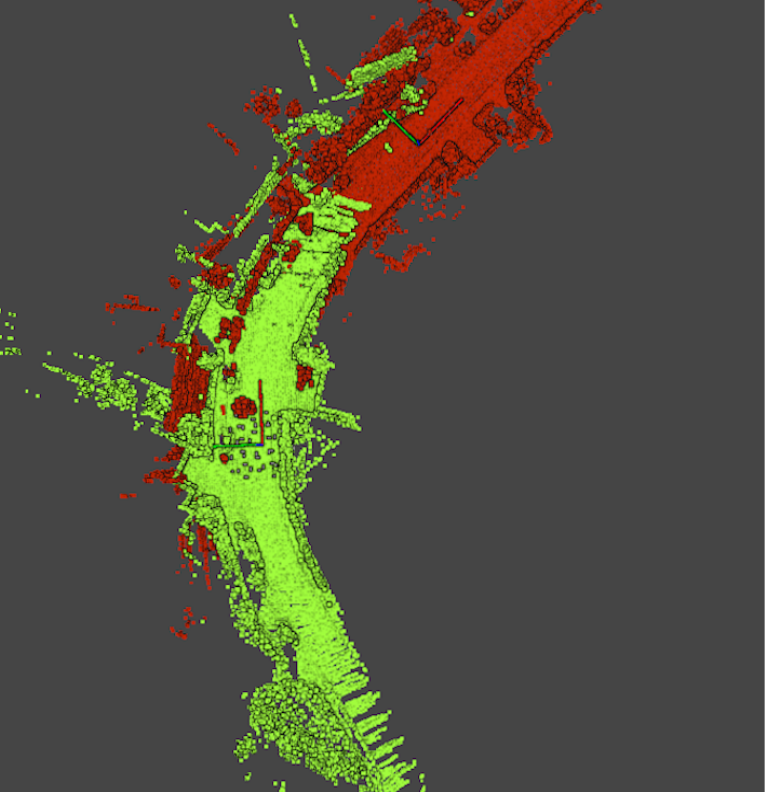}
  \subcaption{Scan matching result}
  \end{minipage}
  \caption{Matching cost factor enables constraint of the relative pose between frames with a small overlap where scan matching can be deteriorated. The orange line in (A) indicates a matching cost factor between two frames, and (B) shows a corrupted scan matching result between these frames.}
  \label{fig:constraint}
\end{figure}

The global mapping module takes the optimized submaps as input and optimizes their poses such that they are all aligned together. Every time a new submap is created, we compare the overlap between that submap and all past submaps, and create a matching cost factor between every submap pair with an overlap rate larger than a small threshold $\text{\it th}^G_{\text{\it min}}$ (e.g., 2.5\%). This results in a densely connected factor graph, as shown in Fig. \ref{fig:kitti00}. The proposed approach aggressively creates matching cost factors between submaps with a very small overlap, where scan matching would fail to align the submaps, and thus obtaining an accurate SE3 relative pose constraint is difficult (see Fig. \ref{fig:constraint}). Although the matching cost factors over such submaps would represent deficient constraints, they do not disrupt the optimization because the entire system is well constrained by other factors. This approach helps in not only implicitly closing loops but also improving the odometry estimation accuracy because every submap is connected to all of the submaps in sight of that submap.

\begin{figure}[t]
  \centering
  \begin{minipage}[b]{0.95\linewidth}
  \centering
  \includegraphics[width=\linewidth]{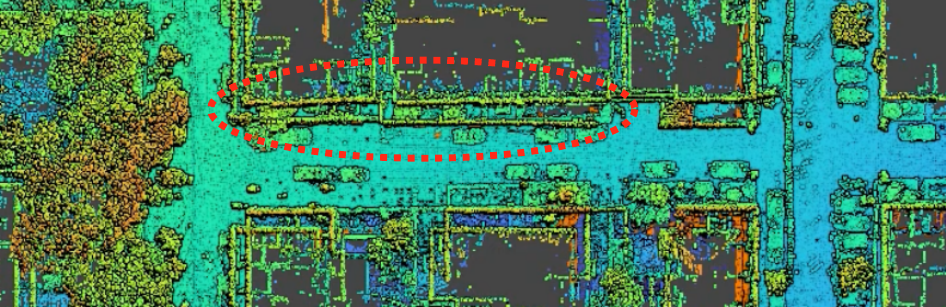}
  \subcaption{Before explicit loop closing}
  \end{minipage}
  \begin{minipage}[b]{0.95\linewidth}
  \centering
  \includegraphics[width=\linewidth]{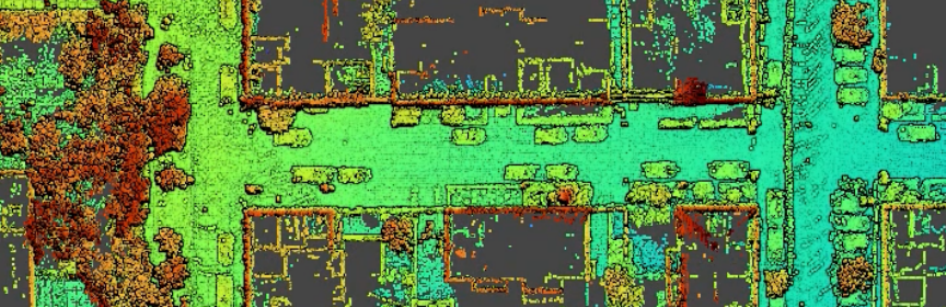}
  \subcaption{After explicit loop closing}
  \end{minipage}
  \caption{Voxel-based matching cost factors can be trapped at a local solution (the region surrounded by the orange oval in (A) has inconsistency). A few loop constraints detected by an external loop detector help the matching cost factors to avoid convergence in a local solution and steer the optimization toward a better solution (B).}
  \label{fig:explicit_loop}
\end{figure}

\begin{figure}[t]
  \centering
  \includegraphics[width=0.95\linewidth]{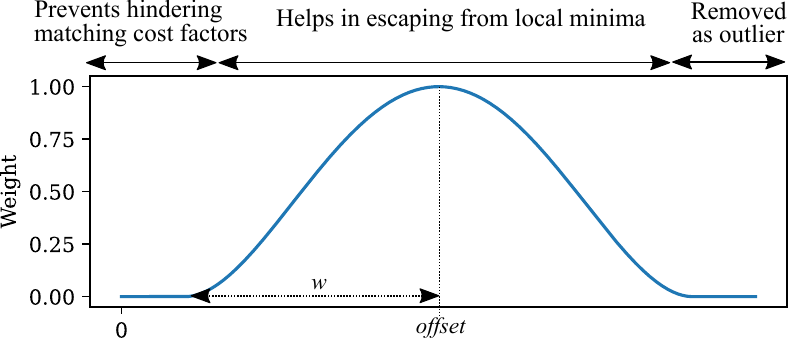}
  \caption{Shifted Tukey's biweight function.}
  \label{fig:shifted_tukey}
\end{figure}

Since the matching cost factor uses voxel-based data association, it can be trapped at a local solution when the estimation drift is large, as shown in Fig. \ref{fig:explicit_loop}(A). In order to overcome this problem, we explicitly detect loops with an external loop detector and add detected loops to the factor graph as SE3 relative pose constraints to help the matching cost factors to escape from local minima. The objective function for the global mapping is thus defined as follows:

\begin{align}
f^{G}(\mathcal{T}) &= f^M(\mathcal{F}^M_G, \mathcal{T}) + f^L(\mathcal{F}^L_G, \mathcal{T}), \\
f^L(\mathcal{F}^L, \mathcal{T}) &= \sum_{(i, j) \in \mathcal{F}^L} \rho \left( \| \log(\hat{\bm{T}}_{ij} ^ {-1} \bm{T}_i^{-1} \bm{T}_j) \|^2 \right),
\end{align}
where $\mathcal{F}^M_G$ is the set of overlapping submap pairs, $\mathcal{F}^L_G$ is the set of loop constraints, $\hat{\bm{T}}_{ij}$ is the relative pose measurement, $\log$ is the logarithmic map, and $\rho$ is a robust kernel. In this work, we obtain explicit loop measurements by applying the conventional GICP to a loop candidate frame pair with initial heading estimate given by ScanContext \cite{Kim2018}.

Although we want the explicit loop constraints to steer the optimization toward a better solution, we want to avoid hindering the matching cost factors when the current estimate sufficiently satisfies the detected loop constraint. For this purpose, we apply Tukey's robust kernel shifted with an offset to each relative pose constraint. The shifted Tukey robust kernel is defined as:
\begin{align}
\label{eq:shifted_tukey}
\mbox{tukey}(x, w) &= \max(0, (1 - x^2/w)^2), \\
\mbox{shifted\_tukey}(x, w, \mbox{\it offset}) &= \mbox{tukey}(\|x - \mbox{\it offset}\|, w),
\end{align}
where $w$ is the kernel width, and $\mbox{\it offset}$ is the amount of shift. As shown in Fig. \ref{fig:shifted_tukey}, this robust kernel forces the optimization in order to satisfy the loop constraint while avoiding disrupting matching cost factors when the relative pose error is small. The kernel also removes loop constraints with errors that are too large as outliers.

With explicit SE3 loop constraints, we aim to steer the optimization toward a better solution but not to correct the trajectory consistency directly, and we need only a few loop detections. We thus use strict loop detection threshold values to avoid false positive loop detections. To build SE3 loop constraints, we simply use a constant covariance matrix. They, however, will not affect the final optimization result because the robust kernel will eliminate them once the current estimate satisfies them. In Fig. \ref{fig:explicit_loop}(B), we can see that the optimization converged in a better solution after adding a few explicit loop constraints.

\subsection {Implementation Details}

For factor graph optimization, we used the Levenberg-Marquardt optimizer in GTSAM\footnote{\url{https://github.com/borglab/gtsam}}. In order to fully leverage GPU acceleration, we used a customized {\it NonlinearFactorGraph} class that first issues all of the cost evaluation tasks on a GPU, performs GPU synchronization, and then collects the calculated results to build a linearized system. Note that we used an efficient reduction technique to compute the summation of Eqs. \ref{eq:cost}, \ref{eq:hessian}, and \ref{eq:coefficients} on a GPU without atomic operations.


\section{Evaluation}

\definecolor{verylightgray}{gray}{0.92}
\newcolumntype{g}{>{\columncolor{verylightgray}}c}

\newcommand{\bfr}{\color{red} \bf}
\newcommand{\bfb}{\color{blue} \bf}
\newcommand{\nabox}{\parbox[c][2mm]{5mm}{\centering -}}
\newcommand{\naboxs}{\parbox[c][2mm]{4mm}{\centering -}}

\begin{table*}[tb]
  \centering
  \caption{Average rotational relative trajectory errors (RTEs) [\textdegree/100m] on the KITTI dataset}
  \label{tab:error_rot}
  \fontsize{7pt}{8pt}\selectfont
  \begin{tabular}{g||g|ggggggggggg|gg}
  \rowcolor{white}
  Sequence Num.                  & Loop      & 00         & 01         & 02         & 03         & 04         & 05         & 06         & 07         & 08         & 09         & 10         & 00-10        & 11-21 \\
  \rowcolor{white}
  Num. of Frames                 & closure   & 4541       & 1101       & 4661       & 801        & 271        & 2761       & 1101       & 1101       & 4071       & 1591       & 1201       & Mean (ST/S) & Mean (ST) \\ \hline \hline

  \rowcolor{white}
  Proposed (matching cost)      &            & 0.16       & \bfb 0.10  & \bfb 0.12  & \bfr 0.19  & \bfb 0.10  & 0.10       & \bfr 0.07  & \bfb 0.11  & 0.18       & \bfb 0.11  & \bfr 0.15  & {\bfb 0.14} / {\bfb 0.13}    & \bfb 0.15 \\ 
  Proposed (matching cost)      & \checkmark & \bfr 0.12  & \bfr 0.09  & \bfr 0.10  & \bfr 0.19  & \bfb 0.10  & \bfr 0.06  & \bfb 0.08  & \bfr 0.10  & \bfr 0.14  & \bfr 0.08  & \bfr 0.15  & {\bfr 0.11} / {\bfr 0.11}    & -         \\
  \rowcolor{white}
  Proposed (SE3)                & \checkmark & 0.18       & 0.15       & 0.17       & 0.33       & 0.17       & 0.21       & 0.10       & 0.17       & 0.50       & 0.17       & 0.31       & 0.24        / 0.22           & -         \\ \hline
  LOAM \cite{Zhang2014}         &            & -          & -          & -          & -          & -          & -          & -          & -          & -          & -          & -          & \nabox      / \naboxs        & \bfr 0.13 \\
  \rowcolor{white}
  MULLS \cite{pan2021mulls}     &            & 0.18       & \bfr 0.09  & 0.17       & \bfb 0.22  & \bfr 0.08  & 0.17       & 0.11       & 0.18       & 0.25       & 0.15       & \bfb 0.19  & \nabox      / 0.16           & 0.19 \\
  MULLS \cite{pan2021mulls}     & \checkmark & \bfb 0.13  & \bfr 0.09  & 0.13       & \bfb 0.22  & \bfr 0.08  & \bfb 0.07  & \bfb 0.08  & \bfb 0.11  & \bfb 0.17  & 0.12       & \bfb 0.19  & \nabox      / {\bfb 0.13}    & -    \\
  \rowcolor{white}
  ELO \cite{elo}                &            & 0.20       & 0.13       & 0.18       & 0.27       & 0.15       & 0.17       & 0.13       & 0.16       & 0.21       & 0.14       & \bfb 0.19  & \nabox      / 0.18           & 0.21 \\
  IMLS-SLAM \cite{Deschaud2018} &            & -          & -          & -          & -          & -          & -          & -          & -          & -          & -          & -          & \nabox      / \naboxs        & 0.18 \\
  \rowcolor{white}
  SuMa \cite{Behley2018}        & \checkmark & 0.23       & 0.54       & 0.48       & 0.50       & 0.27       & 0.20       & 0.30       & 0.54       & 0.38       & 0.22       & 0.32       & 0.36        / 0.36           & 0.34 \\
  SuMa++ \cite{Chen2019}        & \checkmark & 0.22       & 0.46       & 0.37       & 0.46       & 0.26       & 0.20       & 0.21       & 0.19       & 0.35       & 0.23       & 0.28       & 0.29        / 0.29           & 0.34 \\
  \rowcolor{white}
  LiTAMIN2 \cite{yokozuka2021}  & \checkmark & 0.28       & 0.46       & 0.32       & 0.48       & 0.52       & 0.25       & 0.34       & 0.32       & 0.29       & 0.40       & 0.47       & 0.33        / 0.38           & -    \\
  LO-Net \cite{Li2019}          &            & 0.42       & 0.40       & 0.45       & 0.59       & 0.54       & 0.35       & 0.33       & 0.45       & 0.43       & 0.38       & 0.41       & \nabox      / 0.43           & -    \\ 
  \end{tabular}
  \\
  {\bfr Red} and {\bfb blue} respectively indicate the first and second best results. \\
  Mean ST and S respectively indicate the means of sub-trajectory and sequence errors.
\end{table*}

\begin{table*}[tb]
  \centering
  \caption{Average translational relative trajectory errors (RTEs) [m/100m] on the KITTI dataset}
  \label{tab:error_trans}
  \fontsize{7pt}{8pt}\selectfont
  \begin{tabular}{g||g|ggggggggggg|gg}
  \rowcolor{white}
  Sequence Num.                 & Loop       & 00         & 01         & 02         & 03         & 04         & 05         & 06         & 07         & 08         & 09         & 10         & 00-10        & 11-21  \\
  \rowcolor{white}
  Num. of Frames                & closure    & 4541       & 1101       & 4661       & 801        & 271        & 2761       & 1101       & 1101       & 4071       & 1591       & 1201       & Mean (ST/S) & Mean (ST) \\ \hline \hline

  \rowcolor{white}
  Proposed (matching cost)      &            & \bfr 0.49  & 0.65       & \bfr 0.50  & \bfb 0.62  & 0.41       & \bfr 0.24  & \bfb 0.29  & 0.30       & \bfb 0.80  & \bfr 0.46  & \bfb 0.54  & {\bfr 0.52} / {\bfr 0.48}  & \bfb 0.59 \\
  Proposed (matching cost)      & \checkmark & 0.56       & 0.66       & 0.55       & 0.63       & 0.42       & \bfb 0.28  & 0.34       & 0.35       & 0.81       & 0.55       & \bfb 0.54  & 0.56        / 0.52         & -         \\
  \rowcolor{white}
  Proposed (SE3)                & \checkmark & 0.58       & \bfr 0.61  & 0.60       & 0.69       & 0.44       & 0.38       & 0.34       & 0.37       & 1.51       & 0.68       & 0.74       & 0.74        / 0.63         & -         \\ \hline
  LOAM \cite{Zhang2014}         &            & 0.78       & 1.43       & 0.92       & 0.86       & 0.71       & 0.57       & 0.65       & 0.63       & 1.12       & 0.77       & 0.79       & \nabox      / 0.84         & \bfr 0.55 \\
  \rowcolor{white}
  MULLS \cite{pan2021mulls}     &            & 0.51       & \bfb 0.62  & 0.55       & \bfr 0.61  & 0.35       & \bfb 0.28  & \bfr 0.24  & \bfb 0.29  & \bfb 0.80  & 0.49       & 0.61       & \nabox      / {\bfb 0.49}  & 0.65      \\
  MULLS \cite{pan2021mulls}     & \checkmark & 0.54       & 0.62       & 0.69       & \bfr 0.61  & 0.35       & 0.29       & \bfb 0.29  & \bfr 0.27  & 0.83       & 0.51       & 0.61       & \nabox      / 0.52         & -         \\
  \rowcolor{white}
  ELO \cite{elo}                &            & 0.54       & \bfr 0.61  & 0.54       & 0.65       & \bfr 0.32  & 0.33       & 0.30       & 0.31       & \bfr 0.79  & 0.48       & 0.59       & \nabox      / 0.50         & 0.68      \\
  IMLS-SLAM \cite{Deschaud2018} &            & \bfb 0.50  & 0.82       & \bfb 0.53  & 0.68       & \bfb 0.33  & 0.32       & 0.33       & 0.33       & \bfb 0.80  & 0.55       & \bfr 0.53  & {\bfb 0.55} / 0.52         & 0.69      \\
  \rowcolor{white}
  SuMa \cite{Behley2018}        & \checkmark & 0.68       & 1.70       & 1.20       & 0.74       & 0.44       & 0.43       & 0.54       & 0.74       & 1.20       & 0.62       & 0.72       & 0.83        / 0.82         & 1.39      \\
  SuMa++ \cite{Chen2019}        & \checkmark & 0.64       & 1.60       & 1.00       & 0.67       & 0.37       & 0.40       & 0.46       & 0.34       & 1.10       & \bfb 0.47  & 0.66       & 0.70        / 0.70         & 1.06      \\
  \rowcolor{white}
  LiTAMIN2 \cite{yokozuka2021}  & \checkmark & 0.70       & 2.10       & 0.98       & 0.96       & 1.05       & 0.45       & 0.59       & 0.44       & 0.95       & 0.69       & 0.80       & 0.85        / 0.88         & -         \\
  LO-Net \cite{Li2019}          &            & 0.78       & 1.42       & 1.01       & 0.73       & 0.56       & 0.62       & 0.55       & 0.56       & 1.08       & 0.77       & 0.92       & \nabox      / 0.82         & -         \\
  \end{tabular}
  \\
  {\bfr Red} and {\bfb blue} respectively indicate the first and second best results. \\
  Mean ST and S respectively indicate the means of sub-trajectory and sequence errors.
\end{table*}

We evaluated the proposed framework on the KITTI odometry dataset \cite{Geiger2012}. We calculated the relative trajectory errors (RTEs) averaged over 100 to 800 m trajectories with the KITTI official evaluation code ({\it Development kit}). We used an Intel Core i7-8700 (12 threads) with an NVIDIA GeForce GTX 1660 Ti to run the proposed framework. The parameters for the proposed framework used in the evaluation appear on the project page\footnote{See the project page for details: \url{https://staff.aist.go.jp/k.koide/projects/ral2021/index.html}}.


\begin{figure}[tb]
  \centering
  \includegraphics[width=0.9\linewidth]{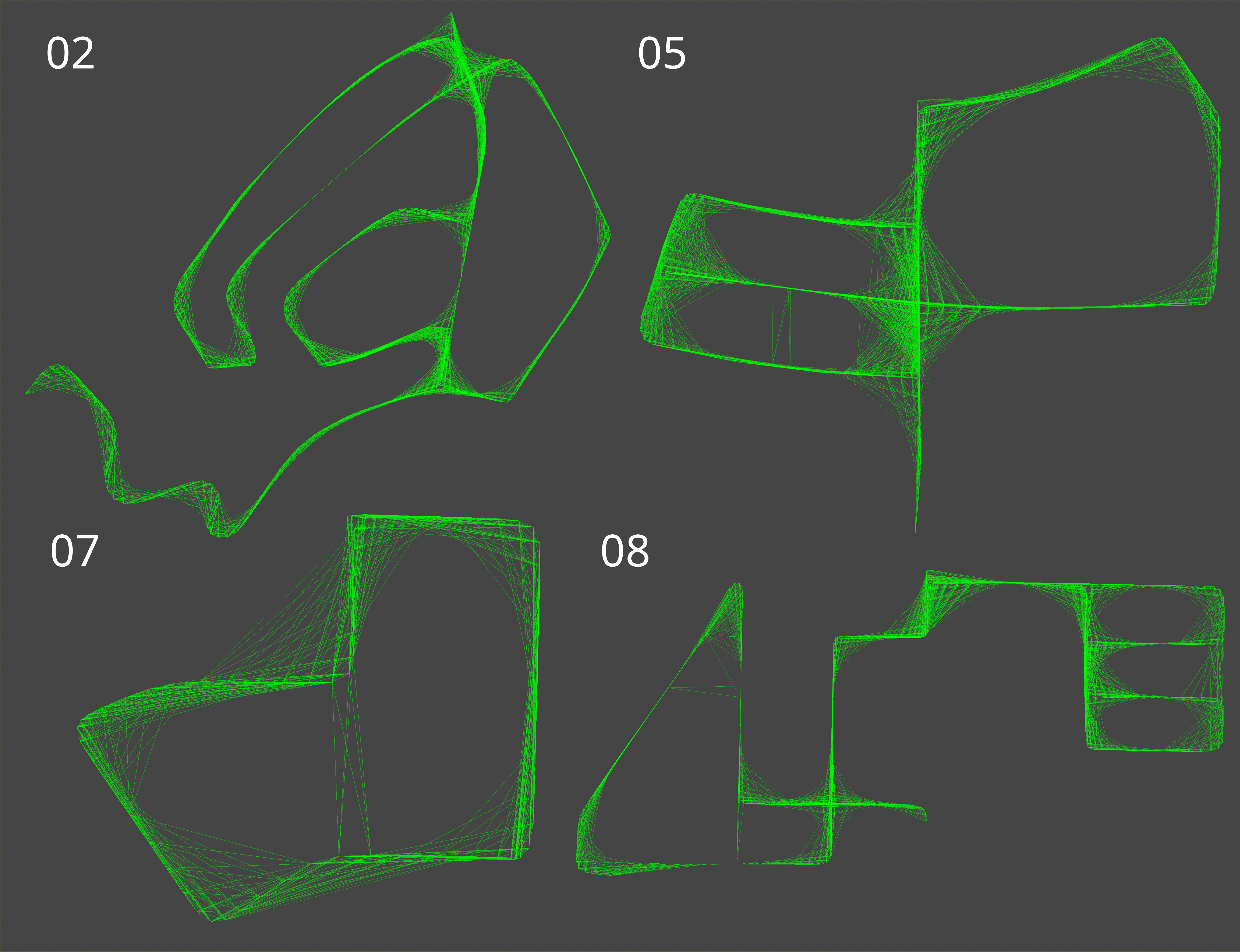}
  \caption{Factor graphs for sequences 02, 05, 07, and 08 generated by the proposed method.}
  \label{fig:graphs}
\end{figure}

{\bf Comparison with State-of-the-art Methods:} We compared the proposed framework with state-of-the-art real-time 3D LiDAR SLAM methods (LOAM \cite{Zhang2016}, MULLS \cite{pan2021mulls}, ELO \cite{elo}, IMLS-SLAM \cite{Deschaud2018}, SuMa \cite{Behley2018}, SuMa++ \cite{Chen2019}, LiTAMIN2 \cite{yokozuka2021}), and a deep-learning-based method (LO-Net \cite{Li2019}). 

We ran the proposed framework with two settings: 1) without implicit and explicit loop closure (i.e., every submap is connected to only other submaps in a sliding window) and 2) with both implicit and explicit loop closure. Similar to \cite{pan2021mulls,elo,Deschaud2018}, we applied an intrinsic vertical scan angle correction to compensate for the point cloud distortion in the KITTI dataset for all the settings.

Tables \ref{tab:error_rot} and \ref{tab:error_trans} show the average rotational and translational RTEs, respectively, of the proposed method and the state-of-the-art methods. We noticed that while the KITTI official benchmark uses the average of sub-trajectory errors to summarize errors, several works report the mean of sequence errors that would overemphasize the errors of short sequences. For a fair comparison, we report both the metrics in Tables \ref{tab:error_rot} and \ref{tab:error_trans} (Means ST: mean of sub-trajectory errors, Mean S: mean of sequence errors).

The proposed method shows the best RTEs (0.14 / 0.13{\textdegree} and 0.52 / 0.48 m) without loop closing among the state-of-the-art methods for the sequence 00 to 10. In particular, the proposed method shows good accuracy on long trajectories (Sequences 00, 02, 05, and 08). For Sequence 11 to 21, the proposed method shows the RTEs that are ranked at the second place among LiDAR-based methods on the KITTI online leaderboard at the time of submission (0.15{\textdegree} and 0.59 m)\footnote{The method {\it GLIM} on \url{http://www.cvlibs.net/datasets/kitti/eval_odometry.php}}. With loop closing, although the rotational RTEs of the proposed method are largely improved, the translational RTEs are slightly deteriorated (0.11 / 0.11{\textdegree} and 0.56 / 0.52 m). Similar trends are reported in several works \cite{Behley2018,pan2021mulls}, and we infer point cloud distortion in the KITTI dataset affected the translational RTEs when loop closing is enabled.

\begin{figure}[tb]
  \centering
  \begin{minipage}[b]{0.44\linewidth}
  \centering
  \includegraphics[height=4.0cm]{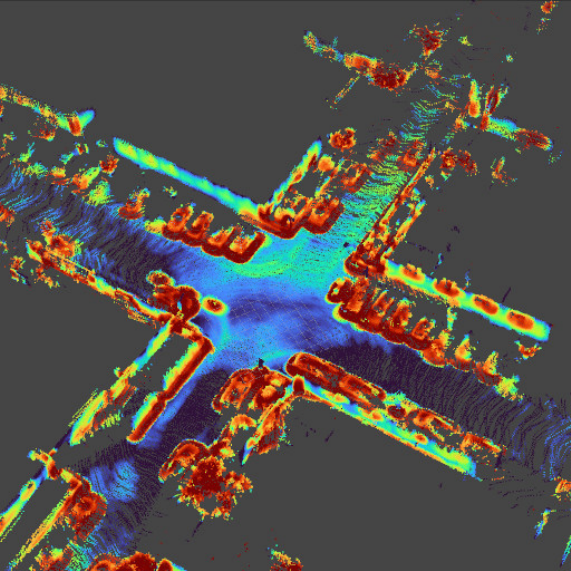}
  \subcaption{SuMa (MME=0.27)}
  \end{minipage}
  \begin{minipage}[b]{0.54\linewidth}
  \centering
  \includegraphics[height=4.0cm]{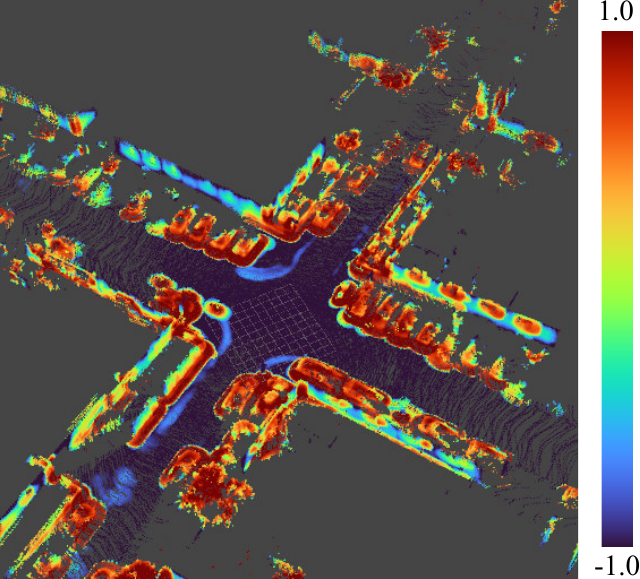}
  \subcaption{Proposed (MME=0.12)}
  \end{minipage}
  \caption{The color indicates per-point entropy (magnitude of inconsistency). SuMa exhibits inconsistent mapping results on the ground and walls due to inaccurate pose graph-based loop closing.}
  \label{fig:mme}
\end{figure}

To assess the mapping quality, we created a local map for every 10 frames by aggregating frames within 10 m and evaluated its mean map entropy (MME) \cite{Razlaw2015}. For the sequence 00, the proposed method showed a small MME (0.14 $\pm$ 0.20) while a pose graph-based method, SuMa \cite{Behley2018}, exhibited a larger MME (0.19 $\pm$ 0.20)\footnotemark[2]. Figure \ref{fig:mme} shows local map MME of the proposed method and SuMa at a junction where a large loop closure happened. We can see points with large entropy (large inconsistency) on the ground and walls of the local map of SuMa, while the proposed method showed significantly smaller entropy (better consistency). This result suggests that the inaccurate modeling of the relative pose constraints in the traditional pose graph optimization can result in inconsistent mapping results while the global matching cost minimization approach can accurately retain the map consistency.


\begin{figure}[tb]
  \centering
  \includegraphics[width=0.9\linewidth]{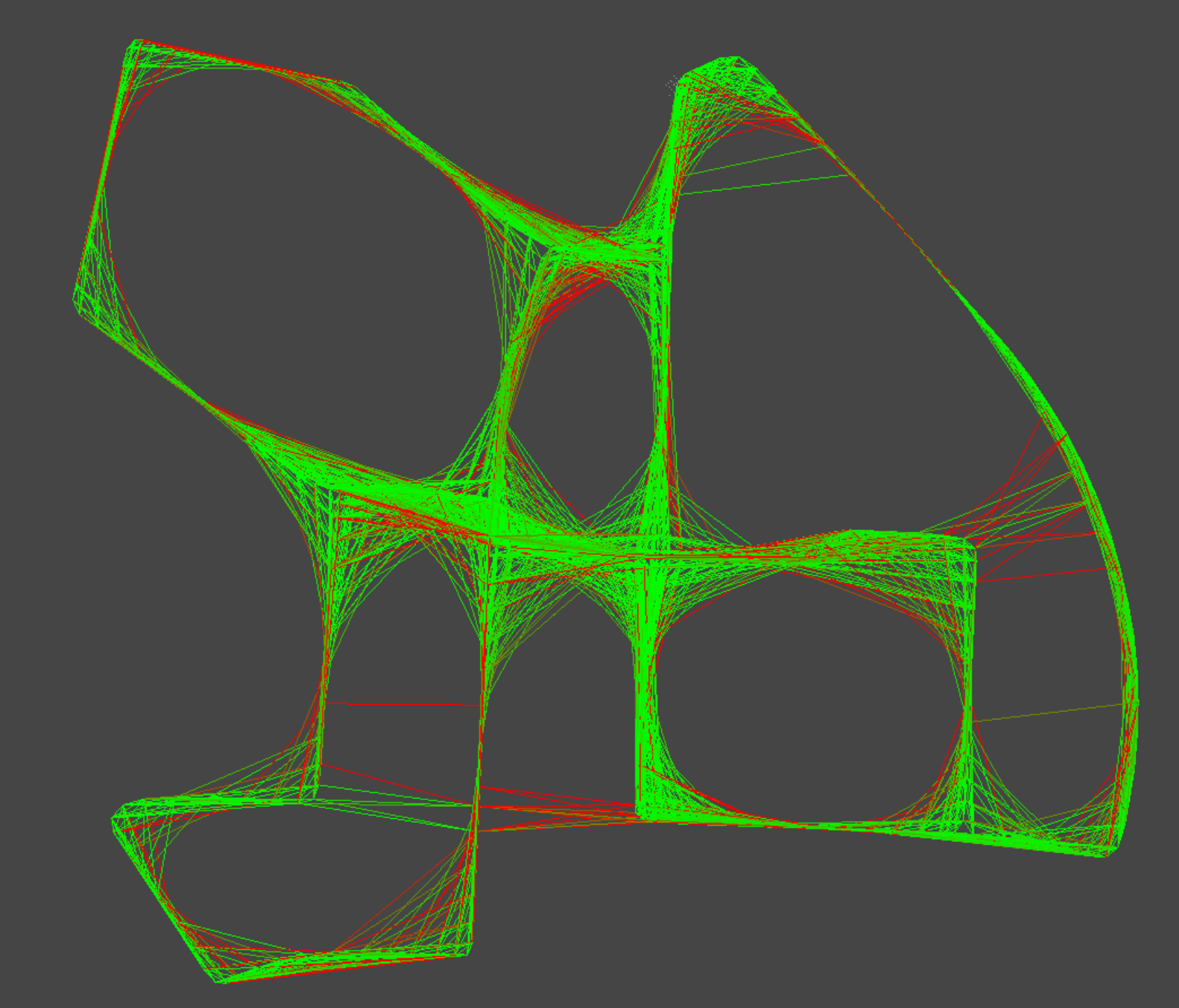}
  \caption{Factor graph with SE3 relative pose factors. The line color indicates the magnitude of the error (Red: large error, Green: small error).}
  \label{fig:se3_graph}
\end{figure}

{\bf Ablation Study:} In order to show that the matching cost minimization enables accurate trajectory estimation in comparison with pose graph optimization, we replaced every matching cost factor with an SE3 relative pose constraint estimated by GICP scan matching \cite{Segal2009}. The initial guess for the scan matching is given based on the optimization result with the matching cost factors. The information matrix of each relative pose factor is calculated based on the Hessian matrix for the GICP scan matching result \cite{Bengtsson2003}. Considering that the scan matching would fail on small overlapping frames, we applied Huber's robust kernel to each relative pose factor.

From Tables \ref{tab:error_rot} and \ref{tab:error_trans}, we can see that the accuracy of the proposed method strongly deteriorated with the relative pose factors, although the graph structure (submap connectivity) had not changed. Figure \ref{fig:se3_graph} shows a factor graph with SE3 relative pose factors. The color of lines indicates the magnitude of errors (Green: small error, Red: large error). We can see that factors between submaps in distance tend to have large errors because the scan matching failed to align the submap pairs with small overlap. The factors with large errors were removed by the robust kernel and thus did not contribute to the optimization result. Note that more factors would have worse relative pose measurements in a practical situation because a good initial guess cannot be expected for scan matching. This result suggests that the pose graph optimization scheme, which requires explicit estimation of the relative pose between frames, has difficulty in constraining distant frames and preserving the consistency over a long trajectory.


\begin{table}[tb]
  \centering
  \caption{Average processing time through KITTI 00}
  \label{tab:proctime}
  \begin{tabular}{c|c|c}
  Module                          & Submodule             & Time [ms]  \\ \hline \hline
  \multirow{2}{*}{Local mapping}  & Factor creation       & 2.8 $\pm$ 5.0 \\
                                  & Optimization          & 123.9 $\pm$ 130.4 \\ \hline
  \multirow{3}{*}{Global mapping} & Factor creation       & 7.7 $\pm$ 12.3 \\ 
                                  & ScanContext           & 7.1 $\pm$ 19.9 \\
                                  & Optimization          & 884.0 $\pm$ 87.6 \\
  \end{tabular}
\end{table}

\begin{figure}[tb]
  \centering
  \includegraphics[width=0.95\linewidth]{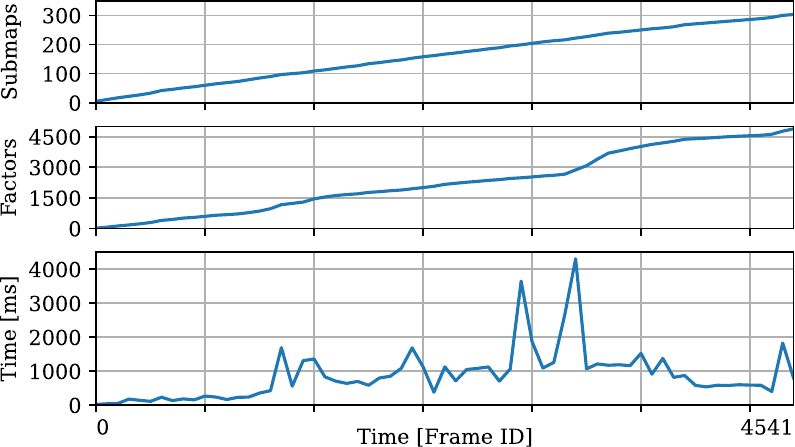}
  \caption{Numbers of submaps and matching cost factors, as well as the global map optimization time for the KITTI 00 sequence.}
  \label{fig:proctime}
\end{figure}

{\bf Runtime:} Through the sequence 00, one of the longest sequences in KITTI, the proposed framework ran approximately twice as fast as the real-time elapsed (20 FPS). Note that we used pre-recorded frontend trajectory estimation results with MULLS \cite{pan2021mulls} (ran at 26 FPS), and thus the processing time of the frontend algorithm was not taken into account. 

Table \ref{tab:proctime} summarizes the runtime of the local and global mapping modules. The local submap optimization, which was performed approximately every 1.5 s, took 123.9 ms on average. The global optimization, which was performed approximately every 7.5 s, took 884.0 ms on average to optimize the factor graph, which had more than 4,500 factors at the end of the sequence by fully leveraging GPU parallel processing. Figure \ref{fig:proctime} shows how the runtime of the global optimization grew as the numbers of submaps and matching factors increased. Although a longer time (3 to 4 s) was required after closing large loops, most of the time, the optimization quickly converged in less than one second. Note that while the linearization and error evaluation of matching cost factors occupied most of the optimization time, the linear solver (performed on a CPU) took only approximately 3\% of the total optimization time on average.

\section{Conclusions} 
\label{sec:conclusion}

This paper presented a 3D LiDAR mapping framework based on VGICP matching cost factors. The GPU-accelerated matching cost evaluation enables simultaneous alignment of all frame pairs in the factor graph and preserves the global consistency over a long trajectory. The local and global mapping modules are managed based on the overlap metric, which can quickly be evaluated on a GPU, and the explicit loop closing mechanism helps the voxel-based matching cost factors to avoid convergence in a local minimum.

\balance

\bibliographystyle{IEEEtran}
\bibliography{ral2021}

\begin{thebibliography}{10}
\providecommand{\url}[1]{#1}
\csname url@rmstyle\endcsname
\providecommand{\newblock}{\relax}
\providecommand{\bibinfo}[2]{#2}
\providecommand\BIBentrySTDinterwordspacing{\spaceskip=0pt\relax}
\providecommand\BIBentryALTinterwordstretchfactor{4}
\providecommand\BIBentryALTinterwordspacing{\spaceskip=\fontdimen2\font plus
\BIBentryALTinterwordstretchfactor\fontdimen3\font minus
  \fontdimen4\font\relax}
\providecommand\BIBforeignlanguage[2]{{%
\expandafter\ifx\csname l@#1\endcsname\relax
\typeout{** WARNING: IEEEtran.bst: No hyphenation pattern has been}%
\typeout{** loaded for the language `#1'. Using the pattern for}%
\typeout{** the default language instead.}%
\else
\language=\csname l@#1\endcsname
\fi
#2}}

\bibitem{Grisetti2010}
G.~Grisetti, R.~Kummerle, C.~Stachniss, and W.~Burgard, ``A tutorial on
  graph-based {SLAM},'' \emph{{IEEE} Intelligent Transportation Systems
  Magazine}, vol.~2, no.~4, pp. 31--43, Dec. 2010.

\bibitem{Behley2018}
J.~Behley and C.~Stachniss, ``Efficient surfel-based {SLAM} using {3D} laser
  range data in urban environments,'' in \emph{Robotics: Science and Systems
  {XIV}}.\hskip 1em plus 0.5em minus 0.4em\relax Robotics: Science and Systems
  Foundation, June 2018.

\bibitem{Reijgwart2020}
V.~Reijgwart, A.~Millane, H.~Oleynikova, R.~Siegwart, C.~Cadena, and J.~Nieto,
  ``Voxgraph: Globally consistent, volumetric mapping using signed distance
  function submaps,'' \emph{{IEEE} Robotics and Automation Letters}, vol.~5,
  no.~1, pp. 227--234, Jan. 2020.

\bibitem{Landry2019}
D.~Landry, F.~Pomerleau, and P.~Giguere, ``{CELLO-3D}: Estimating the
  covariance of {ICP} in the real world,'' in \emph{{IEEE} International
  Conference on Robotics and Automation}.\hskip 1em plus 0.5em minus
  0.4em\relax {IEEE}, May 2019.

\bibitem{liosam2020shan}
T.~Shan, B.~Englot, D.~Meyers, W.~Wang, C.~Ratti, and R.~Daniela, ``{LIO-SAM}:
  Tightly-coupled lidar inertial odometry via smoothing and mapping,'' in
  \emph{{IEEE}/{RSJ} International Conference on Intelligent Robots and
  Systems}, IEEE.\hskip 1em plus 0.5em minus 0.4em\relax {IEEE}, Oct. 2020, pp.
  5135--5142.

\bibitem{Hess2016}
W.~Hess, D.~Kohler, H.~Rapp, and D.~Andor, ``Real-time loop closure in {2D}
  {LIDAR} {SLAM},'' in \emph{{IEEE} International Conference on Robotics and
  Automation}.\hskip 1em plus 0.5em minus 0.4em\relax {IEEE}, May 2016.

\bibitem{Lu1997}
F.~Lu and E.~Milios, ``Globally consistent range scan alignment for environment
  mapping,'' \emph{Autonomous Robots}, vol.~4, no.~4, pp. 333--349, Oct. 1997.

\bibitem{Borrmann2008}
D.~Borrmann, J.~Elseberg, K.~Lingemann, A.~Nüchter, and J.~Hertzberg,
  ``Globally consistent {3D} mapping with scan matching,'' \emph{Robotics and
  Autonomous Systems}, vol.~56, no.~2, pp. 130--142, Feb. 2008.

\bibitem{Sprickerhof2011}
J.~Sprickerhof, A.~Nüchter, K.~Lingemann, and J.~Hertzberg, ``A heuristic loop
  closing technique for large-scale {6D} {SLAM},'' \emph{Automatika}, vol.~52,
  no.~3, pp. 199--222, Jan. 2011.

\bibitem{vgicp}
K.~Koide, M.~Yokozuka, S.~Oishi, and A.~Banno, ``Voxelized {GICP} for fast and
  accurate {3D} point cloud registration,'' in \emph{{IEEE} International
  Conference on Robotics and Automation}.\hskip 1em plus 0.5em minus
  0.4em\relax {IEEE}, May 2021.

\bibitem{Segal2009}
A.~Segal, D.~Haehnel, and S.~Thrun, ``Generalized-{ICP},'' in \emph{Robotics:
  Science and Systems {V}}.\hskip 1em plus 0.5em minus 0.4em\relax Robotics:
  Science and Systems Foundation, June 2009.

\bibitem{Kim2018}
G.~Kim and A.~Kim, ``{Scan Context}: Egocentric spatial descriptor for place
  recognition within {3D} point cloud map,'' in \emph{{IEEE}/{RSJ}
  International Conference on Intelligent Robots and Systems}.\hskip 1em plus
  0.5em minus 0.4em\relax {IEEE}, Oct. 2018.

\bibitem{Geiger2012}
A.~Geiger, P.~Lenz, and R.~Urtasun, ``Are we ready for autonomous driving? the
  {KITTI} vision benchmark suite,'' in \emph{{IEEE}/{CVF} Conference on
  Computer Vision and Pattern Recognition}.\hskip 1em plus 0.5em minus
  0.4em\relax {IEEE}, June 2012.

\bibitem{yokozuka2021}
M.~Yokozuka, K.~Koide, S.~Oishi, and A.~Banno, ``{LiTAMIN2}: Ultra light
  lidar-based slam using geometric approximation applied with
  {KL}-divergence,'' in \emph{{IEEE} International Conference on Robotics and
  Automation}.\hskip 1em plus 0.5em minus 0.4em\relax {IEEE}, May 2021.

\bibitem{pan2021mulls}
Y.~Pan, P.~Xiao, Y.~He, Z.~Shao, and Z.~Li, ``{MULLS}: Versatile lidar slam via
  multi-metric linear least square,'' in \emph{{IEEE} International Conference
  on Robotics and Automation}.\hskip 1em plus 0.5em minus 0.4em\relax IEEE, May
  2021.

\bibitem{Bengtsson2003}
O.~Bengtsson and A.-J. Baerveldt, ``Robot localization based on
  scan-matching{\textemdash}estimating the covariance matrix for the {IDC}
  algorithm,'' \emph{Robotics and Autonomous Systems}, vol.~44, no.~1, pp.
  29--40, July 2003.

\bibitem{Iversen2017}
T.~M. Iversen, A.~G. Buch, and D.~Kraft, ``Prediction of {ICP} pose
  uncertainties using monte carlo simulation with synthetic depth images,'' in
  \emph{{IEEE}/{RSJ} International Conference on Intelligent Robots and
  Systems}.\hskip 1em plus 0.5em minus 0.4em\relax {IEEE}, Sept. 2017.

\bibitem{Whelan2013}
T.~Whelan, M.~Kaess, J.~J. Leonard, and J.~McDonald, ``Deformation-based loop
  closure for large scale dense {RGB-D} {SLAM},'' in \emph{{IEEE}/{RSJ}
  International Conference on Intelligent Robots and Systems}.\hskip 1em plus
  0.5em minus 0.4em\relax {IEEE}, Nov. 2013.

\bibitem{Park2018}
C.~Park, P.~Moghadam, S.~Kim, A.~Elfes, C.~Fookes, and S.~Sridharan, ``Elastic
  {LiDAR} fusion: Dense map-centric continuous-time {SLAM},'' in \emph{{IEEE}
  International Conference on Robotics and Automation}.\hskip 1em plus 0.5em
  minus 0.4em\relax {IEEE}, May 2018.

\bibitem{MurArtal2017}
R.~Mur-Artal and J.~D. Tardos, ``{ORB}-{SLAM}2: An open-source {SLAM} system
  for monocular, stereo, and {RGB-D} cameras,'' \emph{{IEEE} Transactions on
  Robotics}, vol.~33, no.~5, pp. 1255--1262, Oct. 2017.

\bibitem{Schops2019}
T.~Schops, T.~Sattler, and M.~Pollefeys, ``{BAD} {SLAM}: Bundle adjusted direct
  {RGB-D} {SLAM},'' in \emph{{IEEE}/{CVF} Conference on Computer Vision and
  Pattern Recognition}.\hskip 1em plus 0.5em minus 0.4em\relax {IEEE}, June
  2019.

\bibitem{liu2020balm}
Z.~Liu and F.~Zhang, ``{BALM}: Bundle adjustment for lidar mapping,''
  \emph{{IEEE} Robotics and Automation Letters}, vol.~6, no.~2, pp. 3184--3191,
  Apr. 2021.

\bibitem{Wisth2021}
D.~Wisth, M.~Camurri, S.~Das, and M.~Fallon, ``Unified multi-modal landmark
  tracking for tightly coupled lidar-visual-inertial odometry,'' \emph{{IEEE}
  Robotics and Automation Letters}, vol.~6, no.~2, pp. 1004--1011, Apr. 2021.

\bibitem{elo}
X.~Zheng and J.~Zhu, ``Efficient {LiDAR} odometry for autonomous driving,''
  \emph{{IEEE} Robotics and Automation Letters}, pp. 1--1, 2021.

\bibitem{Li2019}
Q.~Li, S.~Chen, C.~Wang, X.~Li, C.~Wen, M.~Cheng, and J.~Li, ``{LO}-net: Deep
  real-time lidar odometry,'' in \emph{{IEEE}/{CVF} Conference on Computer
  Vision and Pattern Recognition}.\hskip 1em plus 0.5em minus 0.4em\relax
  {IEEE}, June 2019.

\bibitem{Chen2020}
X.~Chen, T.~Läbe, A.~Milioto, T.~Röhling, O.~Vysotska, A.~Haag, J.~Behley,
  and C.~Stachniss, ``{OverlapNet}: Loop closing for {LiDAR}-based {SLAM},'' in
  \emph{Robotics: Science and Systems {XVI}}.\hskip 1em plus 0.5em minus
  0.4em\relax Robotics: Science and Systems Foundation, July 2020.

\bibitem{Hu2020}
Q.~Hu, B.~Yang, L.~Xie, S.~Rosa, Y.~Guo, Z.~Wang, N.~Trigoni, and A.~Markham,
  ``{RandLA}-net: Efficient semantic segmentation of large-scale point
  clouds,'' in \emph{{IEEE}/{CVF} Conference on Computer Vision and Pattern
  Recognition}.\hskip 1em plus 0.5em minus 0.4em\relax {IEEE}, June 2020.

\bibitem{Stoyanov2012}
T.~Stoyanov, M.~Magnusson, H.~Andreasson, and A.~J. Lilienthal, ``Fast and
  accurate scan registration through minimization of the distance between
  compact {3D} {NDT} representations,'' \emph{International Journal of Robotics
  Research}, vol.~31, no.~12, pp. 1377--1393, Sept. 2012.

\bibitem{Zhang2014}
J.~Zhang and S.~Singh, ``{LOAM}: Lidar odometry and mapping in real-time,'' in
  \emph{Robotics: Science and Systems {X}}.\hskip 1em plus 0.5em minus
  0.4em\relax Robotics: Science and Systems Foundation, July 2014.

\bibitem{Deschaud2018}
J.-E. Deschaud, ``{IMLS}-{SLAM}: Scan-to-model matching based on 3d data,'' in
  \emph{{IEEE} International Conference on Robotics and Automation}.\hskip 1em
  plus 0.5em minus 0.4em\relax {IEEE}, May 2018.

\bibitem{Chen2019}
X.~Chen, A.~Milioto, E.~Palazzolo, P.~Giguere, J.~Behley, and C.~Stachniss,
  ``{SuMa}++: Efficient {LiDAR}-based semantic {SLAM},'' in \emph{{IEEE}/{RSJ}
  International Conference on Intelligent Robots and Systems}.\hskip 1em plus
  0.5em minus 0.4em\relax {IEEE}, Nov. 2019.

\bibitem{Zhang2016}
J.~Zhang and S.~Singh, ``Low-drift and real-time lidar odometry and mapping,''
  \emph{Autonomous Robots}, vol.~41, no.~2, pp. 401--416, Feb. 2016.

\bibitem{Razlaw2015}
J.~Razlaw, D.~Droeschel, D.~Holz, and S.~Behnke, ``Evaluation of registration
  methods for sparse {3D} laser scans,'' in \emph{European Conference on Mobile
  Robots}.\hskip 1em plus 0.5em minus 0.4em\relax {IEEE}, Sept. 2015.

\end{thebibliography}

\end{document}